\documentclass{article}

\usepackage{arxiv}

\usepackage[utf8]{inputenc} 
\usepackage[T1]{fontenc}    
\usepackage{hyperref}       
\usepackage{url}            
\usepackage{booktabs}       
\usepackage{amsfonts}       
\usepackage{nicefrac}       
\usepackage{microtype}      
\usepackage{lipsum}		
\usepackage{graphicx}
\usepackage{doi}
\usepackage{amsmath}
\usepackage{graphicx}
\usepackage{subcaption}
\usepackage{float}
\usepackage[export]{adjustbox}
\usepackage{array}
\usepackage{amssymb}
\usepackage{multirow}
\usepackage{longtable}
\usepackage{algorithm}
\usepackage{algorithmic}
\usepackage{listings}
\usepackage[dvipsnames]{xcolor}
\usepackage{csquotes}
\usepackage{caption}

\title{Independent Mobility GPT (IDM-GPT): A Self-Supervised Multi-Agent Large Language Model Framework for Customized Traffic Mobility Analysis Using Machine Learning Models}

\author{ \href{https://orcid.org/0009-0000-8734-5498}{\includegraphics[scale=0.06]{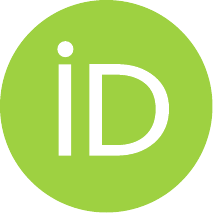}\hspace{1mm}Fengze~Yang} \\
	Department of Civil \& Environmental Engineering\\
	University of Utah\\
	Salt Lake City, 84112 \\
	\texttt{fred.yang@utah.edu} \\
	\And
    \hspace{1mm}Xiaoyue~Cathy~Liu \\
	Department of Civil \& Environmental Engineering\\
	University of Utah\\
	Salt Lake City, 84112 \\
	\texttt{cathy.liu@utah.edu} \\
	\And
    \hspace{1mm}Lingjiu~Lu \\
	Department of Civil and Environmental Engineering\\
    University of Washington,\\
    Seattle, WA, 98195\\
	\texttt{lingjiu@uw.edu} \\
	\And
    \hspace{1mm}Bingzhang~Wang \\
	Department of Civil and Environmental Engineering\\
    University of Washington,\\
    Seattle, WA, 98195\\
	\texttt{bzwang@uw.edu} \\
	\And
	\href{https://orcid.org/0000-0000-0000-0000}{\includegraphics[scale=0.06]{orcid.pdf}\hspace{1mm}Chenxi~(Dylan)~Liu}\thanks{Corresponding author.} \\
	Department of Civil \& Environmental Engineering\\
	University of Utah\\
	Salt Lake City, 84112 \\
	\texttt{chenxi.liu@utah.edu} \\
}

\hypersetup{
pdftitle={Independent Mobility GPT (IDM-GPT): A Self-Supervised Multi-Agent Large Language Model Framework for Customized Traffic Mobility Analysis Using Machine Learning Models},
pdfsubject={},
pdfauthor={Fengze~Yang, Xiaoyue~Cathy~Liu, Lingjiu~Lu, Bingzhang~Wang, Chenxi~Liu},
pdfkeywords={Large Language Models, Traffic Data Analysis, Urban Mobility, Machine Learning},
}

\begin{document}
\maketitle

\begin{abstract}
	With the urbanization process, an increasing number of sensors are being deployed in transportation systems, leading to an explosion of big data. To harness the power of this vast transportation data, various machine learning (ML) and artificial intelligence (AI) methods have been introduced to address numerous transportation challenges. However, these methods often require significant investment in data collection, processing, storage, and the employment of professionals with expertise in transportation and ML. Additionally, privacy issues are a major concern when processing data for real-world traffic control and management. To address these challenges, the research team proposes an innovative Multi-agent framework named Independent Mobility GPT (IDM-GPT) based on large language models (LLMs) for customized traffic analysis, management suggestions, and privacy preservation. IDM-GPT efficiently connects users, transportation databases, and ML models economically. IDM-GPT trains, customizes, and applies various LLM-based AI agents for multiple functions, including user query comprehension, prompts optimization, data analysis, model selection, and performance evaluation and enhancement. With IDM-GPT, users without any background in transportation or ML can efficiently and intuitively obtain data analysis and customized suggestions in near real-time based on their questions. Experimental results demonstrate that IDM-GPT delivers satisfactory performance across multiple traffic-related tasks, providing comprehensive and actionable insights that support effective traffic management and urban mobility improvement.
\end{abstract}


\section{Introduction}
Urbanization drives economic growth, enhances access to services and opportunities, and fosters cultural and social development in concentrated areas. However, it also challenges the equilibrium between infrastructure supply and travel demands, resulting in mobility issues that can slow economic activity due to increased congestion, longer commute times, and higher pollution levels. A report published by the United States Department of Transportation (USDOT) \cite{TSAR2023} reveals that the vehicle miles of travel grew 2.1\% faster than the public road and street mileage between 2010 and 2022, even when accounting for the effects of the pandemic. Concurrently, the annual hours of delay per commuter in "very large" cities rose by 5 hours from 2010 to 2019, prior to the pandemic. These factors collectively diminish living quality and could lead to resident dissatisfaction \cite{GUZMAN2023103765, OLFINDO2021153, MAHESHWARI2024104041}. Therefore, it is significant for transportation agencies like DOTs to address mobility challenges within current transportation systems and enhance network resilience. To achieve this goal, they need to identify the current traffic demands and road system bottlenecks by deploying an extensive array of sensors and devices to collect data, including loop detectors, surveillance cameras, radar or LiDAR sensors, and GPS devices \cite{jiang2024cost, liu2023cooperative}. These sources generate substantial quantities of data, such as speed, traffic volumes, and density \cite{liu2023real}, providing transportation agencies with a detailed understanding of network mobility and supporting the development of more accurate and reliable traffic models. Nevertheless, the surge of big data also presents great opportunities as well as significant challenges, such as processing efficiency, data privacy and security issues, as well as the increased labor costs associated with implementing advanced technologies like Machine Learning (ML) applications.

Leveraging the vast influx of big data, numerous ML techniques have been integrated into transportation systems to capture the temporal-spatial dependencies of traffic data, thereby addressing various mobility challenges. To efficiently harness the potential of traffic data, ML models such as statistical models and deep learning (DL) models offer a broad spectrum of analytical techniques for traffic agencies. Among these, DL models like neural networks (NN) are particularly beneficial for traffic data analysis due to their capabilities in complex pattern recognition, feature extraction, and capturing non-linear spatio-temporal correlations \cite{lv2014traffic, polson2017deep, zhu2018big, liu2020bi, liu2022toward}. For example, Convolutional Neural Networks (CNNs) are well-suited for spatial data analysis and can proficiently process data in grid formats such as images or road network grids; while Recurrent Neural Networks (RNNs), especially Long Short-Term Memory (LSTM) networks, excel at capturing temporal dependencies over sequences of data.  Although these state-of-the-art ML models significantly enhance the utility of traffic data, their direct implementation introduces several challenges:
\begin{itemize}

    \item \textbf{Time Consumption and Efficiency:} Traffic agency analysts take a non-real-time to fetch, process, and analyze large volumes of traffic data. This delay can hinder timely decision-making and real-time response to changing traffic conditions.

    \item \textbf{Scalability and Error Management:} Manual analysis struggles to scale effectively with the vast amounts of traffic data generated by modern sensor deployments. Data overload can lead to potential oversights and errors due to fatigue.
    
    \item \textbf{Costs and Resource Requirements:} Employing human analysts involves significant expenses, including salaries, ongoing training, and infrastructure for manual data collection and processing.

    \item \textbf{Data Privacy:} Data privacy is a critical issue when human analysts manually process traffic data in ML/AI implementations. The manual handling of sensitive information, such as travel routes, timestamps, and personal vehicle data, increases the risk of data breaches or non-compliance with privacy regulations \cite{li2020differential}.
    
\end{itemize}

Nevertheless, the advent of AI agents empowered by large language models (LLMs) mitigates the challenges caused by human analysts applying traditional ML models or neural networks (NNs) \cite{liu2024potentials}. LLMs, such as OpenAI's GPT series, are transformer models that excel in understanding and generating human-like text by learning from vast amounts of textual data, making them highly effective for natural language processing tasks \cite{brown2020language}. In the context of evaluating traffic patterns, AI agents utilize the capabilities of LLMs to function as intermediaries that simplify the interaction with complex ML models, offering intuitive natural language interfaces. These interfaces, brought by LLMs, facilitate ease of use for non-expert users, enabling them to conduct complex queries and interpretations more efficiently, thereby reducing the need for specialized ML expertise.

Moreover, by integrating various ML models with the natural language processing abilities of LLMs, AI agents alleviate the need for traffic agencies to possess extensive ML knowledge or undergo specific training \cite{LIU2023100017}, thus reducing additional labor costs associated with implementing ML models and improving operational efficiency. Additionally, LLM agents contribute to minimizing the risk of data leakage by acting as secure mediators in the processing of sensitive personal travel data, obviating direct human access. Due to their capacity to analyze extensive volumes of unstructured data from diverse sources \cite{rae2021scaling} or to query structured data from databases \cite{rajkumar2022evaluating, zhang2024benchmarking}, LLM agents efficiently identify trends and anomalies, reducing the necessity for manual intervention. By implementing advanced encryption and anonymization techniques during data processing, LLM agents ensure that the underlying data's privacy is preserved while yielding valuable insights.

Although LLM agents show promise for addressing challenges in traffic management, a gap remains between theoretical benefits and practical applications. LLMs require a chain-of-thought (CoT) strategy to break down complex questions for better results \cite{kojima2022large}. For larger tasks like mobility analysis, a single agent struggles to fully address the user queries. Effective solutions require multiple LLM agents collaborating in a structured framework, with effective prompting to improve reasoning and decision-making \cite{zhang2024trafficgpt}. Moreover, LLMs often lack domain-specific knowledge in transportation and ML model implementation \cite{yao2023knowledge}, limiting their effectiveness in real-world scenarios. Therefore, developing a multi-agent LLM framework integrated with ML models tailored for transportation analysis is essential to bridge this gap and enhance their utility for traffic agencies handling complex problems.

Therefore, this paper introduces Independent Mobility GPT (IDM-GPT), a novel multi-agent LLM framework designed to independently support the process of mobility analysis efficiently by utilizing ML models empowered by LLMs. This framework leverages LLMs as an intermediary interface between traffic agencies and complex, high-dimensional traffic data, facilitating a self-organizing, self-optimizing, and self-supervising model for generating traffic analysis results. The framework includes five LLM-based AI agents: Input Validation (IV) Agent, Self-Optimization Prompting (SP) Agent, Database Interaction (DBI) Agent, Data Analysis (DAS) Agent, and Self-Supervision (SS) Agent. Specifically, the IV Agent reviews and validates the user query, extracting pertinent information and defining project objectives. The SP Agent optimizes prompts for better LLM responses. The DBI Agent autonomously retrieves necessary data, enhancing data privacy. The DAS Agent analyzes this data, organizes subtasks with ML models, and generates insights, reducing costs and improving accessibility to advanced techniques. Finally, the SS Agent supervises and optimizes outcomes, ensuring quality results and addressing opacity with clear, actionable insights. The process concludes with the generation of a meticulously crafted report. The contribution lies in:

\begin{itemize}

\item The research introduces an innovative multi-agent LLM framework, IDM-GPT, which autonomously conducts traffic mobility analysis and generates customized traffic management and control suggestions based on user queries in natural language. This system leverages LLMs to efficiently connect transportation users, traffic databases, and ML models.

\item From a methodological perspective, the research proves that the multi-agent LLM framework is able to handle basic traffic mobility tasks with prompting. Additionally, an evaluation LLM is prompted and integrated into IDM-GPT to grade the final output reports, facilitating system optimization. Without fine-tuning and manual evaluation, users can utilize the IDM-GPT easily and reduce the cost of effort and time.

\item IDM-GPT provides user-based analysis reports and customized management suggestions in a privacy-preserving manner. The framework employs LLM Agents to separate users from the database, thereby maintaining the confidentiality of traffic data.

\end{itemize}

Through IDM-GPT, traffic agencies are better equipped to make informed decisions and develop effective emergency traffic control strategies, thereby improving urban mobility management. The structure of the paper is as follows: Section 2 reviews related studies. Section 3 outlines the framework and introduces each AI agent. Section 4 discusses the experimental results on real-world data, evaluates the performance of the framework, and includes an ablation study to demonstrate the effectiveness of the AI agents. Section 5 concludes the paper and suggests directions for future research.

\section{Literature Review}  

The demand to derive traffic movements and patterns from data efficiently in order to ameliorate urban mobility challenges enables traffic agencies to adopt advanced ML methodologies. In this context, LLMs serve as a pivotal tool, potentially enhancing the deployment of ML models in mobility data analysis. This section seeks to encapsulate and evaluate contemporary research on the role of LLMs in two key areas: 1) the challenges associated with these approaches, and 2) the emerging role of LLMs and the multi-agent frameworks in overcoming these challenges. Prior to further discussion on LLMs, this section provides an overview of the current state of research on ML applications in the traffic field, which may serve as a foundational basis for the potential integration of LLMs.

\subsection{ML in mobility analysis}

Traditional statistical models have long been used for traffic forecasting and analysis; however, the non-linear and complex nature of traffic patterns often limits their effectiveness. The advent of ML, particularly deep learning (DL) models, has provided more robust tools for capturing spatiotemporal dependencies in traffic data. Given such nature of traffic data, researchers frequently utilize RNNs, especially LSTM networks, which excel at modeling temporal sequences and have been employed to predict traffic flow based on historical data \cite{ma2015long, fu2016using, yu2017spatiotemporal}. The accuracy of results can be improved by integrating RNNs into various analyses or combining them with other methodologies. For example, combining time series analysis with a 3-layer LSTM-BiLSTM model \cite{ma2021short} and integrating Federated Deep Learning with LSTM \cite{yuan2022fedstn} have yielded promising outcomes. Additionally, some researchers applied attention mechanisms to RNNs, as exemplified by the BiGRU-BiGRU model \cite{chauhan2024novel}, hybrid LSTM, and sequential LSTM \cite{sun2021joint}. These approaches enable the models to capture intricate dependencies within trajectory data, which is crucial for effective network traffic management and congestion mitigation.

To improve the performance of RNN-based models, CNNs have been incorporated into time-series analysis for spatial feature extraction due to their proficiency in handling grid-like data structures, making them suitable for analyzing traffic flow in urban grids \cite{lv2014traffic, zhang2021short}. For instance, the CNN-LSTM hybrid model has been shown to provide accurate predictions by capturing both spatial and temporal traffic features \cite{bogaerts2020graph, zheng2020hybrid}. In another study, LSTM was utilized to capture both inter-day and intra-day patterns, while CNN was employed to integrate contextual information \cite{ma2020daily}.

Graph Neural Networks (GNNs) have also gained attention for modeling traffic networks, representing road networks as graphs to capture the relational information between different nodes (e.g., intersections, road segments) \cite{wu2020connecting}. Recent GNN-related research has focused on addressing the uncertainty of traffic data, thereby enhancing the robustness of GNN models under varying and sparse data conditions. For instance, the Progressive Graph Convolutional Network \cite{shin2024pgcn} learned similarities and extracted temporal features from graph nodes. The Ensemble-Based Spatiotemporal GNN \cite{tanwi2024ensemble} optimized hyperparameter configurations and decomposed uncertainty from GNN ensembles. The Probabilistic GNN \cite{qingyi2024prob} quantified the uncertainty of traffic demand while simultaneously revealing the spatiotemporal pattern of this uncertainty. These models have shown promise in improving the accuracy of traffic prediction and anomaly detection.

Each of these neural network approaches brings distinct advantages to the analysis of traffic data, effectively handling the complex spatial and temporal dependencies within such datasets. However, the wide variety of models in these categories, each with its own strengths and limitations, poses challenges for agencies in terms of scalability, efficiency, the need for expertise in both transportation engineering and ML, as well as concerns about data privacy and security. In contrast, AI agents powered by LLMs could offer a more cost-effective solution, potentially streamlining traffic mobility analysis.

\subsection{AI Agents and LLMs}

The emergence of AI agents empowered by LLMs offers potential solutions to the above-mentioned challenges. AI agents are autonomous systems that interact with their environment through sensors and actuators to make decisions and perform tasks, aiming to achieve specific objectives efficiently and effectively \cite{russell2016artificial}. LLMs, such as OpenAI's GPT series and LLaMA, have demonstrated remarkable capabilities in understanding and generating human-like text \cite{brown2020language}. In transportation, LLMs can serve as intermediaries between users and complex ML models, enabling natural language interfaces, automated data processing, and enhanced decision support. Current LLM-related research in the transportation field can be classified into LLMs as predictors, synthetic data generators, assistants, and evaluators.

\subsubsection{LLM-as-a-Predictor}
LLMs are well-suited for predicting spatial-temporal traffic data due to their ability to capture complex dependencies, integrate multimodal data, and model sequential patterns effectively. The R2T-LLM \cite{guo2024responsiblereliabletrafficflow} was proposed as a novel traffic flow prediction model leveraging LLMs for both accurate and explainable traffic forecasting. However, as noted in the paper, generating coherent and contextually relevant explanations alongside predictions remains a challenge. To help LLMs better understand time-series data, Ren et al. \cite{ren2024tpllm} effectively leveraged pre-trained LLMs for traffic prediction, showing strong performance in both full-sample and few-shot scenarios. However, combining CNNs and GCNs to handle spatiotemporal features increases computational complexity, which may pose challenges for implementation in resource-constrained or real-time systems. Another approach to improving LLMs’ understanding of spatial-temporal data is to tokenize timesteps and locations for embedding in the models. For instance, ST-LLM \cite{liu2024spatial} introduced a partially frozen attention strategy that effectively captures global spatial-temporal dependencies, though its reliance on spatial-temporal embeddings can hinder its suitability for real-time decision-making applications such as dynamic traffic management systems. UrbanGPT \cite{li2024urbangpt} utilized spatiotemporal instruction-tuning to achieve accurate zero-shot predictions in urban scenarios. Although this model is designed for generalization, fine-tuning may still be needed to optimize results for specific urban contexts, which could be challenging for organizations lacking the necessary expertise or computational resources.

\subsubsection{LLM-as-a-Generator}  
LLMs are adept at generating realistic sequence data, making them particularly useful for generating traffic data in simulations. While challenges remain in ensuring the fidelity and diversity of the generated data, LLMs have been widely employed to simulate complex systems such as urban traffic flows and human mobility patterns \cite{long2024llms}. For example, SeGPT, a framework leveraging ChatGPT, excels at generating dynamic and complex scenarios for autonomous vehicle trajectory prediction, aiding in model testing and adaptation across various real-world conditions \cite{li2024chatgpt}. However, generalizing across different driving scenarios, particularly in low-data environments, remains a challenge. Similarly, Chang et al. presented a framework for generating rare and complex corner scenarios for autonomous vehicle testing using LLMs, enhancing scenario diversity and interoperability \cite{chang2024llmscenario}. Nonetheless, the reliance on LLMs to generate complex multi-agent scenarios introduces significant complexity in model design and implementation, which may make it difficult to ensure that the scenarios accurately reflect real-world conditions. MobilityGPT \cite{haydari2024mobilitygpt}, by leveraging a gravity model and road connectivity matrix, successfully generated realistic synthetic trajectories. However, the absence of formal privacy guarantees for the generated data raises concerns when such synthetic data is intended to preserve privacy, potentially exposing sensitive information.

\subsubsection{LLM-as-an-Assistant}  
LLMs can act as powerful assistants, enhancing various aspects of operations, management, and user interaction. Zhang et al. \cite{zhang2024trafficgpt} introduced TrafficGPT, which combines ChatGPT with traffic foundation models (TFMs) to enable natural language interaction with traffic data analysis and decision-making processes. The model translates natural language queries into prompts for TFMs, which then analyze traffic data to produce actionable insights and recommendations. This innovative approach improves the accuracy, efficiency, and reliability of urban traffic management decisions. Similarly, DriveLLM \cite{cui2023drivellm} is a framework that integrates LLMs into autonomous driving systems, enhancing commonsense reasoning and decision-making in dynamic environments. However, in complex environments with multiple moving objects, DriveLLM tends to make overly cautious decisions.

Research on LLM-as-an-Assistant in the transportation field remains relatively underdeveloped compared to their use as predictors or generators. Although some frameworks have explored LLMs for enhancing traffic management and autonomous driving through natural language interaction and decision-making, the scope and depth of this research remain limited. The input methods in these studies may not always be flexible or generic enough, as users might be uncertain about the best approach to achieve traffic project objectives. Additionally, the frameworks require substantial human interaction, which may not always be the most efficient use of time and effort. 

\subsubsection{LLM-as-a-Judge}
To evaluate LLM-generated results, traditional methods like BLEU \cite{papineni2002bleu}, ROUGE \cite{lin2004rouge}, and METEOR \cite{banerjee2005meteor} focus on word and phrase matching but lack semantic and reasoning evaluation. Recent approaches like BERTScore \cite{zhang2019bertscore} and MoverScore \cite{zhao2019moverscore} use contextual embeddings to assess semantics but are sensitive to nuances in word embeddings and may misinterpret out-of-context phrases. To address these issues, LLM-based evaluation methods such as QAGScore, GPTScore, G-Eval, and Prometheus have been developed. These methods offer better semantic understanding and interpretability. QAGScore \cite{fabbri2021qafacteval} evaluates factual consistency through a question-answering approach but is dependent on question-and-answer quality. Prometheus \cite{kim2023prometheus} combines multiple metrics like factuality, fluency, and coherence, but its complex implementation and high computational cost are drawbacks. GPTScore \cite{fu2023gptscore} offers flexibility through fine-tuning for different criteria. Similarly, G-Eval \cite{liu2023g}, based on GPT models, excels at evaluating complex qualities like logical flow and coherence, aligning closely with human judgments. In this paper, though GPTScore is fast and flexible, G-Eval is preferred for its better alignment with human evaluations and logical thinking and reliable and consistent performance in scoring tasks such as coherence, factuality, and relevance.

This paper aims to address the research gap by advancing LLM-as-an-Assistant research and developing a self-motivating, generic framework—IDM-GPT—that assists users in identifying optimal analysis strategies, enhancing data analysis efficiency, decision-making, and user interaction across various transportation tasks.

\section{Methodology}

\begin{figure}
    \centering
    \includegraphics[width=\linewidth]{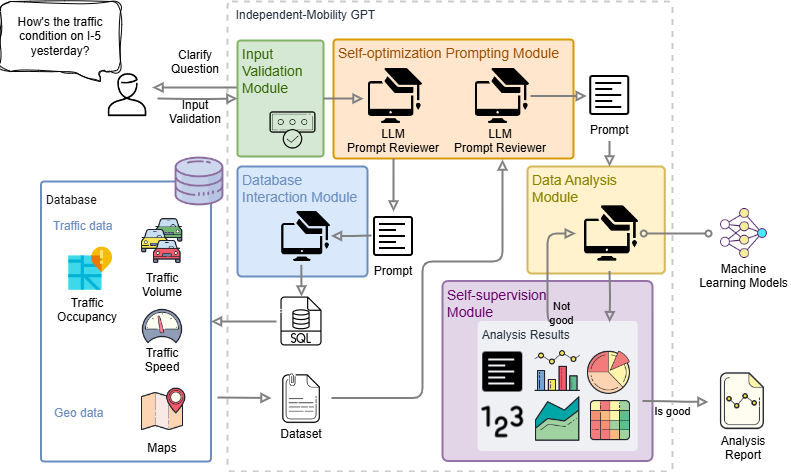}
    \caption{Overview of the IDM-GPT}
    \label{fig:figure1}
\end{figure}

This section elaborates on the detailed mechanisms of how IDM-GPT leverages LLM-based agents to efficiently identify and manage traffic issues. The proposed multi-agent framework integrates advanced ML models with LLMs to process and analyze complex, high-dimensional traffic data. Figure~\ref{fig:figure1} presents the overall process, illustrating how IDM-GPT deconstructs traffic-related tasks through a self-organizing, self-supervising, and self-optimizing approach.

\subsection{Framework overview}
This framework is designed to manage and analyze traffic data using LLMs and ML models, taking user queries as input and producing comprehensive analysis reports that include traffic insights, visualizations, and suggestions. The framework consists of five LLM-based agents:

\textit{IV Agent}: This agent receives user queries and outputs formatted questions with clear objectives and scopes. It prevents irrelevant inputs from wasting computational resources and clarifies the user's intent, time, and location scope to help create better prompts for LLMs, ensuring satisfactory results.

\textit{SP Agent}: This agent takes the validated query or retrieved dataset, drafts a prompt, optimizes it, and forwards it to the next LLM role. It is designed to generate high-performance prompts through iterative optimization.

\textit{DBI Agent}: This agent converts prompts from the previous agent into SQL queries to retrieve necessary data from the database. It ensures that user queries and database descriptions are translated into executable SQL queries effectively.

\textit{DAS Agent}: This agent uses the optimized prompts, including the retrieved datasets, to produce diagrams, analyses, and discussions represented by figures, tables, and text files. It employs various ML models, including NN, to analyze traffic data. Depending on the query, the LLM Data Analyst selects appropriate ML models to identify traffic patterns and reasons for traffic conditions, generating valuable insights for transportation improvements.

\textit{SS Agent}: This agent takes the results from the DAS Agent and produces the final comprehensive analysis report, which includes visualizations and detailed insights. It checks the quality of the analysis results to ensure accuracy and relevance to the user query. Unsatisfactory results are looped back for further refinement.

The entire framework ensures a seamless process from user query to detailed analysis report, leveraging LLMs and ML models for optimal traffic data analysis and insight generation. The examples of prompts of each agent have been attached in Appendix 1.

\begin{figure}
    \centering
    \includegraphics[width=0.9\linewidth]{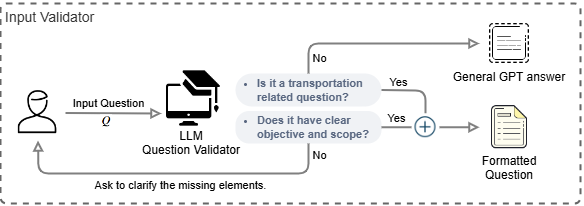}
    \caption{Input validator of the IDM-GPT}
    \label{fig:figure2}
\end{figure}

\subsection{IV Agent}
The process of this agent is illustrated in Figure~\ref{fig:figure2}. This process ensures that user queries are properly framed and relevant to transportation-related issues before further processing. This research firstly uses an LLM-based validator to do a semantic check for the input question \( Q \) by using a predefined vector database \( D \).

Let \( q \) represent the query \( Q \) as a semantic vector and \( q \in \mathbb{R}^n \), where \(\mathbb{R}^n\) is an \( n \)-dimensional vector space, and \( D = \{d_1, d_2, \dots, d_m\} \) represent the set of database entries, where each \( d_i \in \mathbb{R}^n \) is also an \( n \)-dimensional vector. To check the semantic similarity between the query \( q \) and an entry \( d_i \) in the database, this paper uses cosine similarity, which is a common measure for evaluating the similarity between two vectors in semantic space.

The cosine similarity \( S(q, d_i) \) between the query vector \( q \) and the database vector \( d_i \) is given by Eq.~\ref{eq:cos}:
\begin{equation}
    S(q, d_i) = \frac{q \cdot d_i}{\| q \| \| d_i \|}
    \label{eq:cos}
\end{equation}

Where \( \| q \| \) and \( \| d_i \| \) are the Euclidean norms of the vectors \( q \) and \( d_i \). \( S(q, d_i) \in [-1, 1] \), where 1 indicates perfect similarity, 0 indicates no similarity, and -1 indicates complete dissimilarity.

To determine if a query matches semantically with an entry in the database, a threshold \( \tau \) is set as \(S(q, d_i) \geq \tau_d\), where \( \tau_d \) is a pre-defined threshold indicating a sufficient level of semantic similarity for query topic checking.

Next, the Eq.~\ref{eq:Tx} checks if \(Q\) passes the semantic check and is a transportation-related query:
\begin{equation}
    T(X) = 
    \begin{cases} 
    \text{True} &  S(q, d_i) \geq \tau_d \\
    \text{False} & S(q, d_i) < \tau_d
    \end{cases}
    \label{eq:Tx}
\end{equation}

Similarly, a general function \( F \) takes \( q \) and a set of reference vectors \( r_i \in R \), where \( R \) is the database for vectors used to validate the existence of objective and scopes as a reference, as inputs and checks if the similarity of relevant terms or phrases in \( Q \) is greater than threshold \( \tau_r\) :
\begin{equation}
    F(Q, R) = 
    \begin{cases} 
    \text{True} & \text{if } g(q, r_i) \geq \tau_r \\
    \text{False} & \text{if } g(q, r_i) < \tau_r
    \end{cases}
\end{equation}

Where \( R \) can be \( O_{\text{r}} \) for objectives, \( S_{\text{rp}} \) for time scopes, or \( S_{\text{rl}} \) for location scopes.

This formulation allows the representation of conditions as Eq.~\ref{eq:O_Q} to \ref{eq:S_Q}:
\begin{equation}
    O(Q) = F(Q, O_{\text{r}})
    \label{eq:O_Q}
\end{equation}
\begin{equation}
    S_{\text{p}}(Q) = F(Q, S_{\text{rp}})
    \label{eq:S_p}
\end{equation}
\begin{equation}
    S_{\text{l}}(Q) = F(Q, S_{\text{rl}})
    \label{eq:S_Q}
\end{equation}

Finally, the function \( V(Q) \) determines if \( Q \) is valid, ensuring it is transportation-related and contains both a rational objective and scope:

\begin{equation}
    V(Q) = T(X) \land O(Q) \land S_{p}(Q) \land S_{l}(Q)
\end{equation}
where \( \land \) denotes the logical AND operation.

If \(V(Q)\) is true, it is transformed into a formatted question, ready for subsequent steps of data retrieval and analysis. However, if \(V(Q)\) is false, this agent requests clarification from the user to address the missing elements. The user is prompted to provide additional information to ensure that the query is transportation-related and well-defined. For queries that do not meet the criteria even after clarification, the system generates a general GPT response, which may not be specific to transportation.

\subsection{SP Agent}
To seamlessly integrate the user query \(Q\) into system instructions and generate effective LLM prompts, this paper introduces an AI agent that can optimize the prompts automatically. Building on the G-Eval method, this agent evaluates the initial prompt \(P_d\), which contains \(Q\), to identify areas for improvement. The prompt is then iteratively refined until it either reaches an optimal state or hits the maximum refining epoch, as shown in Algorithm~\ref{alg:prompt_opt}. This iterative process guarantees the creation of high-quality prompts, improving the performance of complex LLM tasks within IDM-GPT, such as SQL generation and data analysis.

\begin{algorithm}
\caption{SP Agent}
\label{alg:prompt_opt}
\begin{algorithmic}[1]
\STATE Initialize prompt $P_d$
\STATE Construct enhanced evaluation prompt $P_e$:
\(P_e = P_d + I_e + C_e\)
\STATE SET $epoch \leftarrow 0$
\REPEAT
    \STATE $L_e$ generates chain-of-thought steps $\text{CoT}_e$
    \STATE $P_e$.append($\text{CoT}_e$)
    \STATE Calculate score of $P_d$ using $Score(P_e)$
    \IF{$Score(P_e) < \tau_e$}
        \STATE Update prompt $P_d \leftarrow \mathcal{C}(P_d)$
    \ENDIF
    \STATE INCREMENT epoch
\UNTIL{$Score(P_e) \geq \tau_e$ OR $epoch\ >=\ MAX\ epoch$}
\STATE Return optimized prompt $P_o$
\end{algorithmic}
\end{algorithm}

In this paper, we introduce an enhanced evaluation prompt \(P_e\), which is derived from the initial prompt \(P_d\) by adding two key components: the evaluation instructions \(I_e\) and the evaluation criteria \(C_e\). Specifically, \(P_e\) is expressed as \(P_e = P_d + I_e + C_e\).

The evaluation instructions \(I_e\) guide the evaluator's task, for example:

\textit{You will receive a preliminary prompt. Your task is to refine it for clarity and specificity to improve the quality of the generated response.}

The criteria \(C_e\) outline the dimensions on which \(P_d\) will be assessed, such as clarity, specificity, relevance, and brevity. An example for clarity might read:

\textit{Clarity (1-100) – The overall understandability and straightforwardness of the prompt. A clear prompt minimizes ambiguity and confusion, using simple language and avoiding jargon to ensure the task is precisely conveyed.}

After constructing \(P_e\), the evaluation LLM \(L_e\) generates chain-of-thought \(\text{CoT}_e\) steps to assess \(P_d\). Using the scoring function in Eq.~\ref{eq:scoring}, \(L_e\) then scores \(P_d\) based on both \(P_e\) and \(\text{CoT}_e\).
\begin{equation}
    \label{eq:scoring}
    Score(P_e) = \sum_{i=1}^{n} p(s_i) \times s_i
\end{equation}

Where \(s_i\) represents the score for a given criterion, and \(p(s_i)\) is the probability assigned to that score by \(L_e\). This method allows \(P_d\) to be scored with variance close to human judgments, minimizing the occurrence of ties.

Then a threshold \( \tau_e \) is defined to determine if the similarity score is sufficient. If \( Score(P_e) < \tau_e \), \(P_d\) will be passed through the LLM Prompt Optimizer \(L_{opt}\), which generates optimization suggestions \( \mathcal{C}(P_d) \). The new prompt is then updated:
\begin{equation}
    P_d \leftarrow \mathcal{C}(P_d)
\end{equation}

where \( \mathcal{C} \) is the LLM critique process providing optimization suggestions. The process iterates, refining \( P_d \) until \( Score(P_e) \geq \tau_e \). This iterative process can be formally expressed as Eq.~\ref{eq:P_o}:
\begin{equation}
    P_o = \lim_{n \to \infty} \mathcal{C}^n(P_d)
    \label{eq:P_o}
\end{equation}

Where \( P_o \) is the optimized prompt.

\subsection{DBI Agent}
In this research, system instructions are developed to guide the LLM in generating SQL queries, demonstrating the effectiveness of this approach. To ensure the accurate generation of SQL queries, instructions are meticulously crafted to provide detailed descriptions of the database schema and multiple examples of natural language questions paired with their corresponding SQL queries. These examples are carefully selected to illustrate the pattern and structure of the task, helping the LLM learn how to translate natural language into SQL syntax effectively. To further enhance the model's capability of handling different types of queries and generating appropriate SQL statements, examples covering a wide range of SQL operations, such as selection, filtering, aggregation, joining, and sorting, are incorporated.

The DBI Agent is integral to data privacy management, providing a secure interface between users and traffic databases that contain sensitive trip trajectory information. By mediating access, it minimizes direct human interaction with personal data, significantly reducing the risk of unauthorized exposure. Additionally, the agent employs customized prompts to anonymize personal travel details, ensuring compliance with data protection standards. This proactive approach not only safeguards user privacy but also strengthens data handling practices, fostering trust and reliability in managing sensitive information.

\subsection{DAS Agent}
The DAS Agent in IDM-GPT plays a crucial role in interpreting and visualizing traffic data. This agent takes an optimized prompt as input, which includes a description of the retrieved dataset. The process involves several key steps: understanding the dataset, planning the analysis, selecting appropriate machine learning models, analyzing and visualizing the data, discussing results, and deriving insights and suggestions. This process is represented in Algorithm \ref{alg:data_analysis}.

\begin{algorithm}[H]
\caption{DAS Agent}
\label{alg:data_analysis}
\begin{algorithmic}[1]
\STATE \textbf{Input:} $D$, $Q_{\text{user}}$, $M$
\STATE \textbf{Output:} Analysis result $R$, Insights $I$, Suggestions $S$

\STATE $R, \;I, \;S, \;A \leftarrow$ []

\STATE \textbf{Step 1: Understanding the Dataset}
\STATE $D_{\text{description}} \leftarrow \text{Describe}(D)$
\STATE SET $P = \{D_{\text{description}},\ Q_{\text{user}},\ M\}$

\STATE \textbf{Step 2: Planning the Analysis}
\STATE $A \leftarrow \text{LLM.PlanSteps}(P)$

\FOR{each analysis step $a_i$ in $A$}
    \STATE \textbf{Step 3: Model Selection}
    \STATE $M_j \leftarrow \text{LLM.ModelSelect}(a_i,\ Q_{\text{user}},\ M_j)$
    
    \STATE \textbf{Step 4: Analysis and Visualization}
    \STATE $R_j \leftarrow M_j(D)$

    \STATE \textbf{Step 5: Discussion and Insight Derivation}
    \STATE $I_j \leftarrow \text{LLM.Interpret}(R_j)$
    \STATE $S_j \leftarrow \text{LLM.Suggest}(Q_{\text{user}},\ I_j)$

    \STATE $R$.append($R_j$)
    \STATE $I$.append($I_j$)
    \STATE $S$.append($S_j$)
\ENDFOR

\STATE \textbf{Return} $R, \;I, \;S$
\end{algorithmic}
\end{algorithm}

Algorithm \ref{alg:data_analysis} starts by accepting three inputs: the dataset \( D \), the user query \( Q_{\text{user}} \), and a structured description of alternative ML models \( M \). The expected outputs are analysis results \( R \), insights \( I \) and suggestions \( S \).

The process begins by creating a system instruction \( P \) for the LLM, which includes the dataset description \( D_{\text{description}} \), the user query \( Q_{\text{user}} \), and the available models \( M \). \( D_{\text{description}} \) is generated by a non-LLM function and contains statistical metadata and data types for each column, while \( Q_{\text{user}} \) defines the objective and scope of the query.

The LLM uses \( P \) to generate an analysis plan \( A \), comprising a series of analytical tasks \( a_i \) to address the user’s query. For each task, the LLM selects a suitable model \( M_j \) based on \( D \), \( Q_{\text{user}} \), and the characteristics of each model in \( M \). The selected model \( M_j \) is then applied to \( D \), producing intermediate results \( R_j \).

Next, the LLM interprets \( R_j \) to extract insights \( I_j \) and generate suggestions \( S_j \). The final outputs \( R \), \( I \), and \( S \) are aggregated from all steps, providing a comprehensive set of insights and recommendations addressing the user query.

\subsection{SS Agent}
This agent ensures the quality of traffic analysis results by comparing them against predefined criteria and iteratively optimizing the results if necessary. In this agent, the G-Eval method is employed to obtain the evaluation score \(E\) of the evaluation criteria \(C\). The desired results criteria \( C = \{C_1, C_2,... , C_n\} \) consist of \(n\) key components. Given the initial result \( R \) and the desired result criteria \( C \), the SS Agent operates through a series of steps to ensure the final output meets the specified criteria. This agent refines and improves the LLM's result by allowing the LLM to evaluate and improve its own output using the Recursive Criticism and Improvement (RCI) approach \cite{kim2024language}. RCI is an iterative process designed to refine and enhance prompts to ensure they meet predefined quality standards. The process is shown in Algorithm \ref{alg:supervision}.

\begin{algorithm}
\caption{SS Agent}
\label{alg:supervision}
\begin{algorithmic}[1]
\STATE \textbf{Input:} $R,\ I,\ S$ from \textbf{Algorithm \ref{alg:data_analysis}}, Desired Results Criteria $C$
\STATE \textbf{Output:} Optimized Output Results $R'$, Insights $I'$, Suggestions $S'$

\STATE SET $R' \leftarrow R$
\STATE SET $I' \leftarrow I$
\STATE SET $S' \leftarrow S$
\STATE SET $epoch \leftarrow 0$

\REPEAT
    \STATE $E \leftarrow$ $\mathcal{L}(R',\ I',\ S'\ |\ C)$
    
    \IF{$E$ < $E_{threshold}$}
        \STATE $IS \leftarrow$ $\mathcal{L_{\text{opt}}}(R',\ I',\ S'\ |\ C)$
        \STATE $Prompt \leftarrow$ RCI($R',\ I',\ S', IS$)
        \STATE $R',\ I',\ S' \leftarrow$ LLM.analyze($Prompt$) \COMMENT{\textbf{Algorithm \ref{alg:data_analysis}}}
    \ENDIF
    \STATE INCREMENT epoch
    
\UNTIL{$E\ >=\ E_{\text{threshold}}$ OR $epoch\ >=\ MAX\ epoch$}

\STATE \textbf{return} $R'$, $I'$, $S'$
\end{algorithmic}
\end{algorithm}

In Algorithm \ref{alg:supervision} \( R,\ I,\ S \) is evaluated against \( C \) using an LLM reviewer \( \mathcal{L} \). This evaluation is represented by eq.~\ref{eq:ss1} and eq.~\ref{eq:ss1.5}:

\begin{equation}
    \label{eq:ss1}
    E = \mathcal{L}(R,\ I,\ S\ |\ C)
\end{equation}
\begin{equation}
    E = \sum_{i=1}^{n} p(s_i) \times s_i
    \label{eq:ss1.5}
\end{equation}
where \( E \) denotes the evaluation score on how well the results meet the criteria. The evaluation score \( E \)  is calculated in eq.~\ref{eq:ss1.5}, where \(S=\{s_1, s_2, ..., s_t\}\) represents a set of scores from 1 to \(t\) predefined in the prompt, \(p(s_i)\) represents the probability of \(s_i\) calculated by the LLM. If \( E \) indicates that the results are not satisfactory, i.e. \( E < E_{\text{threshold}}\), the LLM generates improvement suggestions \( IS \) by the optimization process \( \mathcal{L}_{\text{opt}} \). Based on \( IS,\ R'_{\text{old}},\ I',\ \text{and } S\), \( R \) is regenerated by prompting with the RCI approach and calling Algorithm \ref{alg:data_analysis} to produce improved result \( R'_{\text{new}} \). The process iterates until \( E \) meets or exceeds the threshold \( E_{\text{threshold}} \), or the number of recursions reaches the maximum \(MAX\ epoch\). This iterative optimization ensures that the traffic analysis results are continuously improved until they meet the predefined quality standards.

\section{Experiments \& Results Evaluation}

This section aims to validate the performance of the proposed IDM-GPT in real-world transportation applications, especially focusing on the role of LLM in the IV Agent, SP Agent, DBI Agent, DAS Agent, and SS Agent. The data used in the experiment was collected from the loop detectors installed in the Greater Seattle Region. During the experiment, the similarity score between the outputs from IDM-GPT and the analysis results from human operators was used to evaluate the system's performance. Finally, sample results are presented to demonstrate the effectiveness of IDM-GPT.

\subsection{Experiment Configuration}

In this study, GPT-4o serves as the LLM for IDM-LLM, with the system constructed using Python 3.9 and OpenAI APIs. The experiments are conducted on a PC equipped with an NVIDIA RTX 4060 GPU, an Intel Core i9 CPU, and 32 GB of RAM, ensuring a robust computational environment. The choice of GPT-4o is driven by a balance between performance and cost considerations. The ML models provided to IDM-GPT for selection include traditional models and NNs, such as CNNs, LSTMs, and GNNs. For complex real-world tasks conducted by traffic agencies, fine-tuned LLMs can replace GPT-4o, while modified ML models can be employed to achieve better performance, enhanced reliability, and reduced costs.

For the SP Agent, a threshold of \( \tau=0.8 \) is set, and seven criteria are used: 1) clarity and precision, 2) context and relevance, 3) directiveness, 4) appropriate length, 5) structured format, 6) objective and neutral, 7) avoiding ambiguities, and 8) multiple questions. 

For the DBI Agent, this paper defines 7 prompt-SQL categories to guide the LLM in generating SQL queries. These include select, conditional, join, aggregate, filter-with-join, date range, and group by queries, each designed to address specific data needs like retrieving, filtering, or segmenting information. Each category provides a structured prompt-SQL example, enabling the LLM to accurately map user queries to SQL by referencing targeted examples, thus enhancing query precision and effectively meeting user requirements. The example of the prompt-SQL pairs has been attached in Appendix 1 Example of prompt of DBI Agent.

For model selection in the DAS Agent, this paper provides 6 ML models for the LLM to choose from. The models include LSTM for handling long-term temporal patterns and traffic predictions, GNN for modeling and understanding spatial relationships, autoencoders for detecting traffic patterns, random forest for reasoning tasks, reinforcement learning for optimal routing selection, and Hidden Markov Models for detecting traffic flow irregularities. 

For the SS Agent, the evaluation criteria for analysis results are described in the following sub-section. The total evaluation score ranges from \( 0 \) to \( 1 \), with a threshold \( E_{\text{threshold}} \) set at 0.8. The maximum number of epochs is set to 3 for iterative refinement and assessment.

\subsection{Evaluation Criteria}

The LLM results evaluation criteria consist of two components: numerical results evaluation (NRE) and textual results evaluation (TRE). The NRE involves coding algorithms to validate the quality of the data retrieved from databases and to assess the accuracy of numerical analysis results generated by LLMs. The TRE focuses on evaluating the discussions of findings, insights, and suggestions derived from those insights using well-established criteria for LLM output evaluation. These criteria include coherence, relevance, factual accuracy, and logical flow, which are commonly applied in LLM evaluation frameworks such as BLEU, ROUGE, METEOR, and GPTScore. The detailed evaluation framework ensures a comprehensive assessment of both numerical and textual aspects of the LLM outputs, enhancing the reliability and quality of the results.

\begin{itemize}
    \item Data Integrity (DI): Verify that the input data used in the analysis is accurate, complete, and representative of the traffic conditions.
    \item Result Correctness (RC): For queries with ground truth answers, like predictions and incident detections, evaluate the LLM's result in terms of MSE.
\end{itemize}

Given the open-ended nature of the analysis model selection and results of IDM-GPT, the G-Eval method is leveraged for evaluation. The evaluation criteria of the results include:
\begin{itemize}
    \item Model Validation (MV): Ensure the existence of ML model evaluation metrics such as mean absolute error (MAE), root mean square error (RMSE), R-squared, and precision/recall for classification tasks.
    \item Contextual Relevance (CR): Assess whether the insights provided are relevant to the specific traffic issues being analyzed.
    \item Explanation Quality (EQ): Check that the ML model results include clear explanations and interpretations. The rationale behind the predictions should be understandable.
    \item Visualization Clarity (VC): Evaluate the visualizations for clarity, accuracy, and ease of interpretation. Good visualizations should effectively communicate the key insights without being misleading or overly complex.
\end{itemize}
Following the G-Eval method, GPT-4o serves as the evaluation LLM, assessing results based on the above criteria. The evaluator uses the CoT approach to analyze each criterion, generating detailed evaluation steps and assigning a score from 0 (worst) to 1 (best). Probability-based normalization refines these scores for more precise results, capturing the quality of generated texts and aligning with human metrics, offering a robust assessment of the LLM outputs.

\subsection{Data Description}
The database comprises traffic data from over 24,000 loop detectors embedded on 33 major roads in Seattle, WA, including highways and local main streets. The data, collected at a frequency of one minute, includes features such as volume, occupancy, velocity, loop detector location, and loop detector types. This dataset supports various analyses, including traffic volume and speed assessments, traffic conditions evaluations, and traffic pattern analyses, enabling the derivation of valuable insights. In the subsequent experiments, the input query is limited to a time span of one week to manage data costs, computational resources, and processing time effectively.

\subsection{Performance Evaluation}

Based on the database, this paper defines four main evaluation categories: 1) Make Accurate Prediction (MAP), 2) Optimize Operation (OO), 3) Enhance Safety (ES), and 4) Improve Decision-Making (ID-M). Each category is divided into nine subcategories: 1) Real-Time Traffic Prediction (RT-TP), 2) Real-Time Congestion Forecasting (RT-CF), 3) Weather Impact Prediction (WIP), 4) Complex Network Analysis (CNA), 5) Personalized Routing (PR), 6) Incident Detection in Traffic Patterns (IDTP), 7) Accident Risk Factor Identification (ARFI), 8) Predictive Safety Measures (PSM), and 9) Event Impact Scenario Simulation (EISS). With 15 queries per subcategory, a total of 135 queries are designed to evaluate IDM-GPT’s performance in supporting decision-making and improving urban mobility. These queries cover aspects such as traffic prediction, safety, optimization, and simulation. The detailed query structure is provided in \ref{tab:queryStructure1} in Appendix 2.

To benchmark IDM-GPT, this paper uses the vanilla GPT-4o model as a baseline. Since GPT-4o cannot directly access the database, it is first provided with a description of the database and then prompted to generate SQL queries manually. Once the dataset is retrieved, the baseline model is prompted again, along with the dataset and specific queries, to analyze the data and address the user’s queries. This process ensures that the baseline model performs similar analysis steps as IDM-GPT, allowing for a fair comparison between the two models. The evaluation results for both models are shown in Table~\ref{tab:GPTEval} and Table~\ref{tab:IDMEval}.

\begin{table}[H]
    \centering
    \caption{Evaluation Results of Baseline model (GPT-4o)}
    \begin{tabular}{llrrrrrr}
        \hline
        Category & Subcategory & DI & RC & MV & CR & EQ & VC \\
        \hline
        MAP &       RT\_TP & 0.8990 & 0.2044 & - & 0.4966 & - & \textbf{0.3735} \\
          &       RT\_CF & \textbf{0.9593} & 0.2343 & - & 0.4021 & - & 0.2813 \\
          &         WIP & 0.9557 & 0.3182 & - & 0.4313 & - & 0.3702 \\ \hline
          OO &         CNA & 0.9052 & - & - & 0.7504 & - & 0.2589 \\
           &          PR & 0.8126 & \textbf{0.5344} & - & \textbf{0.8510} & - & 0.3211 \\ \hline
          ES &        IDTP & 0.8422 & 0.3592 & - & 0.6085 & - & 0.2949 \\
           &        ARFI & 0.8235 & 0.4096 & - & 0.4548 & - & 0.2900 \\
           &         PSM & 0.7821 & 0.2979 & - & 0.5645 & - & 0.3577 \\ \hline
         IDM &        EISS & 0.7997 & 0.1161 & - & 0.3548 & - & 0.3685 \\
        \hline
    \end{tabular}
    \label{tab:GPTEval}
\end{table}

\begin{table}[H]
    \centering
    \caption{Evaluation Results of IDM-GPT}
    \begin{tabular}{llrrrrrr} \hline 
    Category & Subcategory &     DI &     RC &     MV &     CR &     EQ &     VC \\ \hline 
             MAP &       RT\_TP & 0.8851 & 0.5409 & 0.9057 & 0.7943 & 0.8793 & 0.6979 \\  
             &       RT\_CF & \textbf{0.9245} & 0.3638 & 0.9870 & 0.8577 & 0.9479 & \textbf{0.8998}\\  
             &         WIP & 0.8561 & 0.3459 & 0.8369 & 0.7295 & 0.8311 & 0.7848 \\ \hline 
              OO &         CNA & 0.7629 & - & 0.8121 & 0.7672 & 0.7961 & 0.7327 \\  
              &          PR & 0.8287 & 0.7296 & 0.9391 & 0.9257 & 0.9615 & 0.7700 \\ \hline 
              ES &        IDTP & 0.8221 & 0.5713 & \textbf{0.9839} & 0.9178 & \textbf{0.9980}& 0.8752 \\  
              &        ARFI & 0.8799 & \textbf{0.8147}& 0.9390 & 0.9592 & 0.9719 & 0.8491 \\  
              &         PSM & 0.8198 & 0.6094 & 0.9413 & \textbf{0.9872}& 0.9460 & 0.7373 \\ \hline 
             ID-M &        EISS & 0.6987 & 0.4189 & 0.8245 & 0.6621 & 0.8311 & 0.7608 \\ \hline
    \end{tabular}
    \label{tab:IDMEval}
\end{table}

The value for the subcategory CNA under the RC criterion is blank in both tables because reasoning questions in this category are open-ended, allowing for multiple correct answers. As such, no ground truth is established for these queries.

In Table~\ref{tab:GPTEval}, the columns MV and EQ are empty because the baseline model cannot automatically leverage machine learning models without explicit prompting. Without ML-driven analysis, the baseline model cannot derive sufficient insights from the dataset, resulting in lower evaluation scores for user queries in other criteria as well.

For the DI criterion, both IDM-GPT and the baseline GPT models perform similarly with scores on the same scale. This is because the data retrieval step follows the same prompting strategy in both cases, ensuring that the input data used for analysis is of comparable quality. Thus, the consistency in data input leads to similar results in terms of data integrity for both models.

During the experiment, the baseline GPT-4o struggled to visualize complex data accurately without detailed prompting. Although the diagrams generated by the baseline model were well-designed and included all necessary elements like axis labels, legends, and titles, the numerical values it selected were incorrect. As a result, while the titles and visual objectives appeared meaningful, the overall VC scores were significantly lower than those of IDM-GPT.

\begin{figure}
    \centering
    \includegraphics[width=0.65\linewidth]{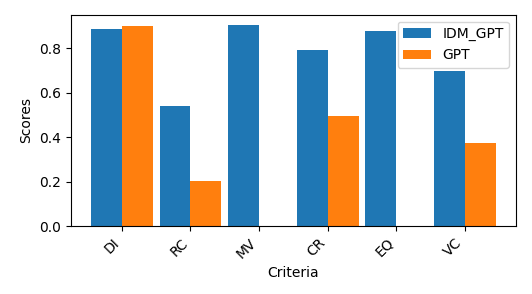}
    \caption{Average Scores for Each Criterion}
    \label{fig:eval_by_criteria}
\end{figure}

The DI scores suggest that IDM-GPT's DBI Agent is more effective in identifying and retrieving congestion-related datasets and relevant forecasting data. Although both models have similar DI scores, the baseline GPT-4o scores are significantly lower on RC, highlighting its struggle to correctly address mobility-related queries, even when given adequate data. This shortcoming negatively impacts its scores in CR and VC, as its analytical performance is insufficient for detailed insights. In contrast, IDM-GPT leverages ML models to generate deeper insights, enabling it to address user queries more effectively and achieve high scores across the remaining criteria.

\begin{figure}
    \centering
    \includegraphics[width=0.65\linewidth]{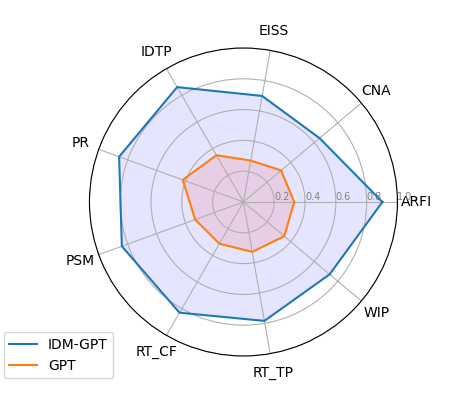}
    \caption{Average Scores for Each Subcategory of Query}
    \label{fig:eval_by_query}
\end{figure}

Based on Figure~\ref{fig:eval_by_query}, IDM-GPT excels in ES queries, showcasing its strong ability to reason about traffic safety factors. Its performance is slightly lower on MAP queries due to limitations in the prediction accuracy of its ML models, though it remains robust overall. In contrast, IDM-GPT faces challenges with OO and ID-M queries, as optimizing traffic networks involves complex, experience-driven adjustments and interdisciplinary considerations. This indicates that integrating more specialized knowledge could enhance its analysis of OO queries. For ID-M queries, performance is constrained by the current ML models in use. Introducing advanced models, such as state-of-the-art diffusion models or generative adversarial networks, could significantly improve IDM-GPT’s capability to generate synthetic data, suggesting a direction for future enhancements.

\subsection{Ablation Study}
IDM-GPT consists of five agents, with the DBI Agent and DAS Agent being essential to the framework's functionality, while the other three agents can be omitted. To validate the effectiveness of the IV Agent, SP Agent, and SS Agent, an ablation study was conducted by repeating the experiment and sequentially omitting each agent. The results of the average evaluation scores are compared among different agent-skipped systems and the baseline system without any agent skipped. The evaluation is summarized in Table \ref{tab:ablation}.

\begin{table}
    \centering
    \caption{Result of ablation study. Bold numbers indicate the numbers with the largest difference compared to the baseline system without skipping any agents.}
    \label{tab:ablation}
    \begin{tabular}{>{\raggedleft\arraybackslash}p{0.1\linewidth}>{\raggedleft\arraybackslash}p{0.1\linewidth}r>{\raggedleft\arraybackslash}p{0.1\linewidth}>{\raggedleft\arraybackslash}p{0.1\linewidth}>{\raggedleft\arraybackslash}p{0.1\linewidth}>{\raggedleft\arraybackslash}p{0.1\linewidth}>{\raggedleft\arraybackslash}p{0.1\linewidth}} 
     \hline
    \textbf{Agent Skipped} & \textbf{DI}  &\textbf{RC}& \textbf{MV} & \textbf{CR} & \textbf{EQ} & \textbf{VC} & \textbf{Avg}\\ \hline
    IV                      & \textbf{0.5234}&0.4052& 0.8125      & 0.6473      & 0.8322& 0.6138& \textbf{0.6391}\\ 
    SP                      & 0.8985&0.5326& 0.9275& 0.7740& 0.8092& \textbf{0.6240}& 0.7610\\ 
    SS                      & 0.9154&0.4792& 0.8084& 0.7265      & \textbf{0.6919}& 0.5913      & 0.7021\\ 
    None                    & 0.9264       &0.5724& 0.9714& 0.7987& 0.9012& 0.7065& 0.8127\\  \hline
    \end{tabular}

\end{table}

When the IV Agent is omitted, the framework performs poorly on the DI, RC, and CR scores. Without reformatting the user query or extracting its objective and scope, the LLM-based DBI Agent struggles to understand the user’s intent, resulting in inadequate dataset retrieval. Consequently, the DAS Agent fails to generate relevant insights, leading to incorrect and ineffective responses to the user query.

The absence of the SP Agent lowers performance across all criteria. Without prompt optimization, the prompts become unclear, diminishing the effectiveness of LLM-based agents like DBI and DAS. Sub-optimally structured prompts cause IDM-GPT to overlook important requirements, resulting in inaccurate analyses and incomplete visualizations, which hinders its ability to derive meaningful insights.

Omitting the SS Agent primarily impacts all criteria except DI, as it operates in the later stages of IDM-GPT. While data retrieval remains relatively unaffected, the absence of SS compromises the final analysis. Without SS, the generated findings lack the necessary model validation metrics and clear, simplified visualizations, both of which are crucial for easy interpretation. This leads to sub-optimal insights that do not fully address the user query.

Overall, the average value shows each agent's importance to IDM-GPT. The lower the value, the more important the agent is. These findings emphasize the essential roles that the IV, SP, and SS Agents play in ensuring the accuracy, relevance, and quality of IDM-GPT's outputs. Each component is critical to maintaining the framework’s performance and delivering meaningful results.

\section{Conclusions}

Acquiring relevant traffic data and selecting appropriate models to support traffic management can be costly, time-consuming, and labor-intensive for transportation agencies. The complexity of advanced ML models demands significant resources for accurate results, while the widespread use of data collection devices raises concerns over personal mobility data privacy. IDM-GPT, as an AI agent, addresses these challenges by streamlining traffic data analysis, enabling the efficient and accurate processing of high-dimensional, spatio-temporal traffic data. Additionally, the AI agent ensures data privacy is maintained throughout the process, mitigating risks associated with handling sensitive personal mobility information. This research makes several notable contributions:

\begin{itemize}
    \item Framework Connecting Traffic Agencies to Databases via LLMs: IDM-GPT effectively bridges traffic agencies with complex traffic data systems, simplifying the interaction by interpreting user queries and generating SQL queries for efficient data retrieval and analysis.
    \item Enhancing Accessibility and Reducing Costs: By integrating LLMs with ML models, IDM-GPT improves usability across various traffic scenarios. The framework’s ability to select and apply appropriate ML models, such as CNNs, LSTMs, and GNNs, reduces both costs and the level of expertise required while demonstrating accuracy in recognizing and predicting traffic patterns.
    \item Improving Data Privacy: IDM-GPT enhances data privacy by leveraging LLMs to access remote databases without direct human intervention. Users only receive the final report, ensuring secure data processing and maintaining the confidentiality of personal traffic data.
\end{itemize}

Despite these contributions, this study has limitations. Reliance on online LLMs, such as ChatGPT, raises stability concerns, suggesting the need for offline models to ensure consistent performance. Additionally, the model selection process could be improved by refining the application of ML models in specific circumstances and expanding the range of models to include more cutting-edge options. Furthermore, the evaluation process could be optimized by finding a better combination of criteria weights rather than averaging evaluation scores to determine the final result.

Future research will focus on developing domain-specific AI agents tailored for various transportation analysis tasks, incorporating advanced multi-modal data processing. Further exploration could also examine real-time data integration and applying IDM-GPT across diverse urban environments. Integrating IDM-GPT with emerging technologies like edge computing and IoT devices offers potential for further improvements.

In conclusion, this research demonstrates the effectiveness of combining LLMs with ML models for traffic data analysis. IDM-GPT provides a scalable solution for improving traffic management and urban mobility, offering a significant contribution to the field. It lays a strong foundation for future advancements in transportation research and practical applications.

\section{Acknowledgments}
The authors express their gratitude to the Washington State Department of Transportation for providing traffic loop detector data and their invaluable support for this research. Additionally, ChatGPT-4o was utilized for polishing the manuscript.

\section{Author Contributions}
Fengze Yang contributed to the conceptualization, methodology, software development, and original draft preparation. Xiaoyue Cathy Liu contributed to the writing (review and editing). Lingjiu Lu contributed to the literature review and experiments. Bingzhang Wang was involved in data investigation. Chenxi (Dylan) Liu (corresponding author) was responsible for project administration supervision and contributed to the writing (review and editing). 

\section{Declaration of Conflicting Interests}
Xiaoyue Cathy Liu is a member of the Transportation Research Record's Editorial Board.

All other authors declared no potential conflicts of interest with respect to the research, authorship, and/or publication of this article.

\section{Funding}
The authors disclosed no financial support for the research, authorship, and/or publication of this article.

\newpage
\bibliographystyle{IEEEtran}
\bibliography{references}

\newpage
\appendix
\section{Appendix 1:}

\subsection{Example of prompt of IV Agent}
\begin{displayquote}
\textit{1. Semantic Inspection for Topic Validity}  

   \begin{itemize}
       \item \textit{Goal: Verify that the user query concerns transportation or mobility analysis. This validation involves checking the semantic similarity between user-provided terms and keywords or phrases in the predefined vector database (e.g., "traffic flow," "transit planning," "mobility pattern," "road safety," "public transport efficiency" <other example phases>).}
       \item \textit{Instruction: Compare user query keywords with the vector database examples. If the similarity score exceeds the predefined threshold, consider the query as a topic-valid question for transportation.}
   \end{itemize}

\textit{2. Detailed Inspection for Objective and Scope Clarity}  

    \begin{itemize}
        \item \textit{Goal: Confirm that the user query has a clear, specific objective and scope, ensuring it’s well-defined for further data retrieval or analysis.}
        \item \textit{Instruction: Check whether the query contains explicit objectives (e.g., "optimize traffic signals," "analyze commuter patterns," "assess public transit impact" <other example phases>) and scope (e.g., "specific locations, timeframes, datasets" <other example phases>) that frame the question in actionable terms. If these conditions are met, classify the query as sufficiently defined.}
    \end{itemize}
\end{displayquote}

\subsection{Example of prompt of SP Agent}
\begin{displayquote}
Step 1: Chain-of-Thought (CoT) Prompt Generation

Generate a step-by-step Chain-of-Thought (CoT) reasoning to assess the quality of this prompt. Consider each aspect carefully:
\begin{itemize}
    \item Relevance: Does the prompt accurately focus on the transportation analysis or mobility task it targets?
    \item Clarity: Is the prompt expressed clearly, so that it is easy to interpret and respond to?
    \item Completeness: Does the prompt contain all necessary information, such as objectives, scope, and context?
\end{itemize}
Provide a detailed CoT that analyzes these elements before giving your final assessment.

Step 2: Prompt Evaluation with CoT Integration

Using the Chain-of-Thought analysis generated, evaluate the prompt on a scale of 1 to 5, based on:
\begin{itemize}
    \item Relevance to Task: Whether the prompt is focused on the intended task.
    \item Clarity of Language: How easily the prompt can be understood.
    \item Completeness: If the prompt includes sufficient details for the LLM to respond effectively.
\end{itemize}
Provide a score for each criterion (1-5), followed by a brief justification for each score. If any score falls below 4, indicate that further improvements are needed and proceed to generate specific suggestions.

Step 3: Prompt for Improvement Suggestions

Based on the evaluation scores, identify one or two specific areas where this prompt could be improved to enhance clarity, relevance, or completeness. For each suggested improvement, provide a brief explanation and an example of how to apply the suggestion to make the prompt more effective. Focus on making the prompt more suitable for the transportation analysis or mobility task.
\end{displayquote}

\subsection{Example of prompt of DBI Agent}
\begin{displayquote}
Part 1: Database Schema Illustration

Here is the schema for the database you will be interacting with. Each table includes a description of its columns, data types, and any relationships between tables. Use this schema to understand the database structure and identify relevant tables and fields for the query.

Database Schema:

$Table 1: `<Table_Name>`
    Column 1: `<Column\_Name>` - `<Data\_Type>` - `<Description>`
    Column 2: `<Column\_Name>` - `<Data\_Type>` - `<Description>`
    <Relationship to other tables or specific constraints, if any>$

$Table 2: `<Table_Name>`
    Column 1: `<Column\_Name>` - `<Data\_Type>` - `<Description>`
    Column 2: `<Column\_Name>` - `<Data\_Type>` - `<Description>`
    <Relationship to other tables or specific constraints, if any>$

<Include schema details for each table as needed>

Use this schema to identify relevant tables and columns for retrieving the information requested in the user query.

Part 2: SQL Query Examples

Here are some example SQL queries that demonstrate typical patterns and syntax for querying this database. Refer to these examples when writing your query.

Example SQL Queries:

1. Basic Select Query: Retrieve basic information from a single table.
   - Example Prompt: "What are the recorded traffic volume, speed, and density data from the traffic monitoring sensors?"
   - SQL:
     \begin{lstlisting}[language=SQL]
     SELECT * FROM [DatabaseX].[Table1];
     \end{lstlisting}

2. Conditional Query: Retrieve specific rows using a WHERE clause.
   - Example Prompt: "What are the traffic volume and speed details for northbound traffic?"
   - SQL:
     \begin{lstlisting}[language=SQL]
     SELECT [Volume], [Speed] 
     FROM [DatabaseX].[Table1] 
     WHERE [Direction] = 'N';
     \end{lstlisting}

3. Join Query: Fetch data across related tables.
   - Example Prompt: "What is the traffic volume and speed data recorded at each sensor location?"
   - SQL:
     \begin{lstlisting}[language=SQL]
     SELECT t1.[Location], t2.[Volume], t2.[Speed] 
     FROM [DatabaseX].[Table1] AS t1 
     JOIN [DatabaseX].[Table2] AS t2 
     ON t1.[ID] = t2.[ID];
     \end{lstlisting}

4. Aggregate Function: Summarize data using aggregate functions.
   - Example Prompt: "What is the average speed across all monitored locations?"
   - SQL:
     \begin{lstlisting}[language=SQL]
     SELECT AVG([Speed]) AS AverageSpeed
     FROM [DatabaseX].[Table3];
     \end{lstlisting}

5. Filtering with Join: Combine join and conditional filtering.
   - Example Prompt: "What are the traffic volumes recorded by sensors within the milepost range of Segment ID 1?"
   - SQL:
     \begin{lstlisting}[language=SQL]
     SELECT t1.[Volume] 
     FROM [DatabaseX].[Table1] AS t1 
     JOIN [DatabaseX].[Table4] AS t4 
     ON t1.[Route] = t4.[Route] 
     AND t1.[Direction] = t4.[Direction]
     WHERE t4.[SegmentID] = 1 AND t1.[Milepost] 
     BETWEEN t4.[MilepostStart] AND t4.[MilepostEnd];
     \end{lstlisting}

6. Date Range Query: Filter based on a date range.
   - Example Prompt: "What is the average traffic speed data for January 2023?"
   - SQL:
     \begin{lstlisting}[language=SQL]
     SELECT AVG([Speed]) AS AverageSpeed 
     FROM [DatabaseX].[Table5]
     WHERE [Date] 
     BETWEEN '2023-01-01' AND '2023-01-31';
     \end{lstlisting}

7. Group By with Aggregation: Group data by a specific column and apply aggregation.
   - Example Prompt: "What is the total traffic volume for each route, and which routes have the highest traffic density?"
   - SQL:
     \begin{lstlisting}[language=SQL]
     SELECT [Route], SUM([Volume]) AS TotalVolume 
     FROM [DatabaseX].[Table6]
     GROUP BY [Route];
     \end{lstlisting}

Use these as templates to ensure query structure aligns with standard SQL practices and strip of any personally identifiable information.

Part 3: SQL Query Generation

Now, based on the database schema provided and examples above, write an SQL query to retrieve data relevant to the user's objective and scope as specified in their query. Ensure the query accurately targets the user’s specified data, including any constraints, filters, or joins needed to match the objective and scope.

User Query: `<Insert the objective and scope of user query>`

Write an SQL query that meets these requirements, ensuring accuracy in selecting tables, columns, conditions, and any necessary joins to retrieve the appropriate data.
\end{displayquote}

\subsection{Example of prompt of DAS Agent}

\begin{displayquote}
    The dataset covers a period of 6 months and includes \\
    observations for 100 detectors located along I-5 highways.
    
    The dataset includes numerical columns <col\_name\_1>, \\
    <col\_name\_2>, ...; includes categorical columns <col\_name\_3>, 
    ...

    <col\_name\_1> \\
    <statistical\_information>

    ...

    <col\_name\_3> \\
    <categorical\_value>

    ...

    Now you should <objective\_in\_user\_query> <scope\_in\_user\_query>.

    Select machine learning models from the following JSON.\\
    \{\\
    \ <model\_name>: \{\\
    \ \ <features>: <natural\_language\_description>, \\
    \ \ <application\_scenarios>: <natural\_language\_description>, \\
    \ \ <usage\_steps>: <natural\_language\_description> \\
    \ \}, \\
    \ ... \\
    \}
\end{displayquote}

\newpage
\section{Appendix 2:}

\begin{table}[H]
    \centering
\caption{Experimental query structure}
\label{tab:queryStructure1}
    \begin{tabular}{>{\raggedright\arraybackslash} m{9em} >{\raggedright\arraybackslash}m{15em} >{\raggedright\arraybackslash}m{10em}} \hline  
         Category&  Sub-category&  Query Example\\ \hline  
         Make Accurate Prediction&  Real-Time Traffic Prediction Forecasting &  Forecast real-time speed changes at SR-520 during road closure period on next weekend.\\ &&\\
             & Real-Time Congestion Forecasting & Estimate congestion spread on I-90 from Bellevue to Seattle during Friday evening peak hours using current occupancy data.\\ &&\\
             & Weather Impact Prediction&Last winter, SR-99 was icy on <date1>, <date2>, ..., <dateN>. Predict occupancy variations on SR-99 during icy road conditions forecasted for tomorrow morning.\\ \hline
         Optimize Operation&  Complex Network Analysis&  Optimize lane utilization on I-405 southbound during heavy traffic periods, balancing speed, occupancy, and vehicle throughput.\\ &&\\
             & Personalized Routing&Suggest optimal routes for commuters traveling from Bellevue to Downtown Seattle, prioritizing minimum travel time and avoiding congested segments.\\ \hline 
         Enhance Safety&  Incident Detection in Traffic Patterns&  Detect sudden drops in speed on I-5 near downtown Seattle to identify potential accidents during morning rush hours.\\ \hline
    \end{tabular}
\end{table}

\begin{table}[H]
    \centering
\label{tab:queryStructure2}
    \begin{tabular}{>{\raggedright\arraybackslash} m{9em} >{\raggedright\arraybackslash}m{15em} >{\raggedright\arraybackslash}m{10em}} \hline 
         Category&  Sub-category&  Query Example\\ \hline  
         Enhance Safety& Accident Risk Factor Identification& The weather for the past season was <weather data>. Identify critical factors contributing to increased accident risks on SR-520 bridge last season.\\ &&\\ 
             & Predictive Safety Measures&I-5 from <milepost1> to <milepost2> was icy on <dates> for past 3 years. Forecast hazardous zones on I-5 during icy conditions next winter to optimize de-icing operations\\ \hline 
         Improve Decision-Making& Event Impact Scenario Simulation& Here are the multi-vehicle accident locations and time periods on I-5, <data>. Simulate the impact of a multi-vehicle accident on I-5 near Northgate to evaluate emergency response strategies.\\ \hline
    \end{tabular}
\end{table}
\end{document}